\documentclass[11pt]{article}

\usepackage{ranlp}
\usepackage{epsf}
\usepackage{amssymb}
\usepackage{graphicx}

\title{Adapting a general parser to a sublanguage}

\author{Sophie Aubin*, Adeline Nazarenko* 
        \and
        Claire N\'edellec**\\
       (*) LIPN, University of Paris 13 \& CNRS UMR 7030\\
	99, av. J.B. Cl\'ement, F-93430 Villetaneuse, France\\
    \texttt{\{sophie.aubin,nazarenko\} at lipn.univ-paris13.fr}\\
          (**) Unit\'e Math\'ematique Informatique et G\'enome (MIG, INRA)\\
    Domaine de Vilvert, F-78350 Jouy en Josas Cedex, France\\
       \texttt{claire.nedellec at jouy.inra.fr}}

\begin{document}

\maketitle

\begin{abstract}
In this paper, we propose a method to adapt a general parser (Link Parser) to sublanguages, focusing on the parsing of texts in biology. Our main proposal is the use of terminology (identification and analysis of terms) in order to reduce the complexity of the text to be parsed. Several other strategies are explored and finally combined among which text normalization, lexicon and morpho-guessing module extensions and  grammar rules adaptation. We compare the parsing results before and after these adaptations.   
\end{abstract}

\section{Introduction}
Most available NLP tools are developed for general language while processing technical texts, {\em i.e.} sublanguages, becomes a necessity for various applications like extracting information from biological texts 
(see \cite{grishman01},\cite{PYY04}, \cite{Grover04} and \cite{Tsu05}). In order to assist the biologists in their daily bibliographical work, the ExtraPloDocs project\footnote{ExtraPloDocs website : http://www-lipn.univ-paris13.fr/RCLN/Extra/ExtraPloDocs/\\These results are also exploited for the development of specialized search engines in the ALVIS project (STREP)~: http://cosco.hiit.fi/search/alvis.html} develops the natural language processing and machine learning tools that enable to build focused information extraction systems in genomics (gene-protein interaction, gene fonctionalities, gene homologies, etc.) at a reasonable cost.  Beyond keyword and statistics based approaches, extracting such relational information must be based on syntax to achieve good precision and coverage (see for instance \cite{Din03}). We therefore need a reliable syntactic parsing of the texts dealing with genomics. 

Instead of redeveloping new parsers for each sublanguage, we try to define a method for adapting a general parser to a specific sublanguage. This paper presents a strategy to adapt the Link Parser (LP) \cite{ST91} to parse Medline abstracts dealing with genomics.

In this paper, we first discuss the question of sublanguages and the different strategies that can be adopted to parse technical texts. Section \ref{cont} presents the context of the adaptation of the LP to the biological domain. In section \ref{diag}, we analyse several cases of parsing failure along with the solutions we propose to adapt the parser. We finally present the evaluation of the modifications we made on the LP grammar and lexicon.

\section{Previous works}\label{Pw}
Sublanguages have been studied for a long time even though it remains a rather confidential part of linguistic and NLP studies. It is noticeable that in specific domains of knowledge, among certain communities and in particular types of texts, people have their own way of writing.  These specific languages are called either sublanguages \cite{harris89,grishman&kittredge86}, restricted or specialized languages depending on the fact that one focuses on the continuity or the gap between these languages and the ``usual language''. 
 In fact, a sublanguage is a restricted (fewer lexicon items and semantic classes) as well as a deviant language (original lexicon items and phrasings). 
This is also noticeable from a distributional point of view. As Harris noticed it, a sublanguage can be characterized by its selectional restrictions and more generally by the distribution of lexicon items and syntactic patterns. 

\cite{sekine97} has argued that parsing should be domain dependent. Three alternative approaches can be considered. Several NLP teams have decided to develop a specialized parser for a given sublanguage (see for instance the String project \cite{sager&al87} or \cite{pust02})  but this approach is considered too expensive for many applications.  A second track consists in training a grammar from a specialized corpus, which requires annotated corpora that are rare in specialized domains. 
An intermediate approach aims at manually adapting a parser as proposed in \cite{PYY04}. This is our approach. 
This work can be considered as a preliminary work to evaluate the potentialities of automating this adaptation.

Two different approaches have been explored for the parsing evaluation. The first is linguistically oriented and based on test suites, a set of sentences that illustrates the various syntactic structures that a parser is supposed to analyse like in TSNLP \cite{lehman&al96}.
The second approach, more pragmatic and more common, consists in evaluating the performances of a parser on a given corpus supposed to be representative of the textual data to parse. 
We will show in the following that we adopted a mixed approach.

As we will see below, one of the main problems in parsing sublanguages is the ambiguity of prepositional attachment.

\section{Context}\label{cont}
 \subsection{The corpora}\label{corpus}
Three different corpora were built from Medline\footnote{http://www.ncbi.nlm.nih.gov/entrez/query.fcgi} abstracts (in English) dealing with transcription in \textit{Bacillus subtilis}. 
As recommended by  \cite{PS00} and \cite{SSD98}, we mixed the two evaluation standards by randomly selecting 212 sentences that we organized according to their linguistic specificities. Despite its relatively small size, the \begin{small}\texttt{MED-TEST}\end{small} corpus is a good sample of the sublanguage of genomics. We also used a larger corpus of full abstracts (\begin{small}\texttt{TRANSCRIPT}\end{small}, 16,981 sentences, 434,886 words) and the \begin{small}\texttt{GIEC}\end{small} corpus made of 160 sentences expressing gene/protein interactions. The \begin{small}\texttt{GIEC}\end{small} corpus was built and used as a benchmark corpus in the context of the Genic Interaction Extraction Challenge\footnote{http://genome.jouy.inra.fr/texte/LLLchallenge} joint to the ICML 2005.




\subsection{The initial parser choice}
In the context of our IE task, and particularly for the ontology acquisition, we need reliable and precise syntactic relations between the words of the whole sentence (except empty words). For those reasons, a symbolic dependency-based parser seemed to be the most adequate.

LP presents several advantages among which the robustness, the good quality of the parsing, the adequation of the dependency technique and representation with our IE task and the declarative format of its lexicon. From the results of the evaluation that we did on different parsers with the \begin{small}\texttt{MED-TEST}\end{small} corpus, it turned out that dependency-based parsers have better results on long and complex sentences, particularly with coordinations. This conclusion is shared by \cite{Din03} who also worked on Medline abstracts. Other experiments, in the context of the ExtrAns project \cite{mol00a}, showed that 76\% of 2,781 sentences from a Unix manpage corpus were completely parsed by LP with no regard to the parsing quality, while we reach only 54\% on the biological corpus. When looking at the quality of the parses, we noticed different kinds of errors depending either on the biological domain or on more general linguistic difficulties like ambiguous constructions. We propose three solutions to address these issues, the text normalization, the use of terminology and the adaptation of the lexicon/grammar of LP.

\section{Diagnosis and adaptation}\label {diag}
  
Our analysis of the performance of the Link grammar on the biological corpus confirms previous works. The main problems can be classified along the following axes.
\subsection{"Textual noise"}
Scientific texts present particularities that we chose to handle in a normalization step prior to the parsing. First, the segmentation in sentences and words was taken off from the parser and enriched with  named entities recognition and rules specific to the biological domain. We also delete some extratextual information that alter the parsing quality. Finally, we use dictionaries and transducers to replace genes and species names by two codes, which prevents from extending the LP dictionary too much.


\subsection{Unknown words}
In the \begin{small}\texttt{TRANSCRIPT}\end{small} corpus, we identified 6,005 out-of-lexicon forms (45,804 occurences) among 12,584 distinct words, {\em i.e.} 47.72\%. They are mostly latin words, numbers, DNA sequences, gene names, misspellings and technical lexicon.

However, LP includes a module that can assign a syntactic category to an unknown word. It is based on the word suffix. Modifying the morpho-guessing (MG) module seemed a better strategy than extending the dictionary since biological objects differ from an organism to another. We then created 19 new MG classes for nouns (\textit{-ase}, \textit{-ity}, etc.) and adjectives (\textit{-al}, \textit{-ous}, etc.) along with their rule.

In the same time, we added about 500 words of the biological domain to the LP lexicon in different classes, mainly nouns, adjectives and verbs. 



\subsection{Specific constructions}
Some words already defined in the LP lexicon present a specific usage in biological texts, which implied some modifications including moving words from one class to another and adaptating or creating rules.

The main motivation for moving words from one class to another is that the abstracts are written by non-native English speakers. This point was also 
raised by \cite{PYY04}. One way to allow the parsing of such ungrammatical sentences is to relax constraints by moving some words from the countable to the mass-countable class for instance. 

Some very frequent words present idiosyncratic uses (particular valency of verbs for instance), which induced the modification or creation of rules.
Numbers and measure units are omnipresent in the corpus and were not necessarily well described or even present in the lexicon/grammar. 
Other minor changes were made that are not mentioned in this paper.
\subsection{Structural ambiguity}\label{amb}
We identified two cases of ambiguity that can be partially resolved by using terminology.
 
\begin{figure}
\begin{center}
\includegraphics[scale = 0.6]{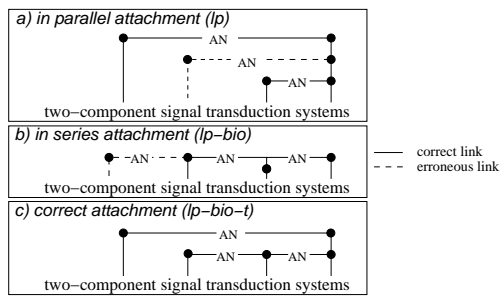}
\end{center}\caption{Series of nouns dependencies}\label{fig1}
\end{figure}

Prepositional attachment is a tricky point that is often fixed using statistical information from the text itself \cite{hindle&rooth93,Fabre01}, a larger corpus \cite{Bou04}, the web \cite{volk02,GALA03} or an external resources such as WordNet \cite{stetina&nagao97}. 
The second major ambiguity factor is the attachment of series of more than two nouns. 
As shown in Figure \ref{fig1}, neither a parallel attachment (lp) nor a serial one (lp-bio) seem to be satisfying.
 We noticed that such cases often appear inside larger nominal phrases often corresponding to domain specific terms. For this reason, we decided to identify terms in a pre-processing step and to reduce them to their syntactic head. If needed, the internal analysis of terms is added to the parsing result for the simplified sentence (see lp-bio-t). The strategy proposed by \cite{Sut95} that consists in the linkage of the words contained in a compound (for instance \textit{``sporulation\_process''}) was excluded. It makes the lexicon size augment and does not reduce complexity for reasons due to the implementation of LP. 

\begin{figure*}
\begin{center}
\includegraphics[scale = 0.6]{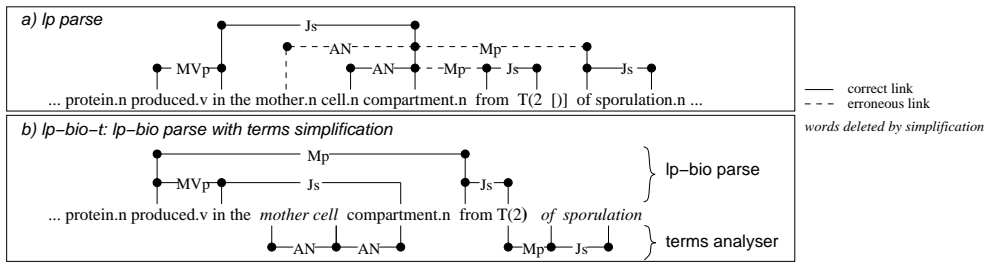}
\end{center}\caption{Example of parsing}\label{fig2}
\end{figure*}

Figure \ref{fig2} shows the influence of the adaptation on the parsing with the fixing of a segmentation error and the disambiguation of prepositional and nominal attachements. 

Before practically integrating the use of terminology in our processing suite, we made a simulation of this simplification of terms. 

\section{Evaluation}
We performed a two-stage evaluation of the modifications in order to measure the respective contribution of the LP adaptation on the one hand and of the term simplification on the other hand.
\subsection{Corpus and criteria}
We used a subset (10 files\footnote{141 sentences, 2630 words}) of the \begin{small}\texttt{MED-TEST}\end{small} corpus but, contrary to the first evaluation (choice of a parser), we wanted to look at the quality of the whole parse and not only to specific relations. 

Table \ref{lex} (for the \begin{small}\texttt{MED-TEST}\end{small} subset) 
shows the way that out-of-lexicon words (OoL), i.e. unknown (UW) and guessed (GW) words, are handled by giving the percentage of incorrect morpho-syntactic category assignations with the original resources (lp), those adapted to biology (lp-bio) and finally the latter associated with the simplification of terms (lp-bio-t).

\begin{table}
\begin{center}
\begin{tabular}{|c||c|c||c|c||c|c|}
\hline
 & \multicolumn{2}{|c||}{\footnotesize{lp}} &\multicolumn{2}{|c||}{\footnotesize{lp-bio}} &  \multicolumn{2}{|c|}{\footnotesize{lp-bio-t}}\\
\hline
& \scriptsize{a} & \scriptsize{b} & \scriptsize{a} & \scriptsize{b} & \scriptsize{a} & \scriptsize{b}\\  
\hline
\hline
\footnotesize{UW} &\scriptsize{244} & \scriptsize{41.4\%} & \scriptsize{53} &\scriptsize{52.8\%} & \scriptsize{26} & \scriptsize{19.2\%}\\
\hline
\footnotesize{GW} &\scriptsize{24} &\scriptsize{4.2\%} & \scriptsize{72} & \scriptsize{0\%} & \scriptsize{31} & \scriptsize{0\%} \\
\hline
\footnotesize{OoL} &\scriptsize{268} & \scriptsize{38\%} & \scriptsize{125} & \scriptsize{22.4\%} & \scriptsize{57} & \scriptsize{8.8\%}\\
\hline
\multicolumn{7}{l}{\scriptsize{a~: total MS assignations, b~: \% of incorrect assignations}}
\end{tabular}\caption{Incorrect MS category assignations} \label{lex}
\end{center}
\end{table}

In Table \ref{res}, five criteria inform on the parsing time and quality for each sentence~: the number of linkages (NbL), the parsing time (PT) in seconds, the fact that a complete linkage is  found or not (CLF), the number of erroneous links (EL) and the quality of the constituency parse (CQ). (NbW) is the average number of words  in a sentence which  varies with the term simplification. The results are given for each one of the three versions of the parser. 

UW, GW, NbL, PT and CLF are objective data while EL and CQ necessitate a linguistic expertise. The CQ evaluation consisted in the assignation of a general quality score to the sentence.

\subsection{Results and comments}

The \textbf{extension of the MG module} reduced the number of erroneous morpho-syntactic category assignations (see Table \ref{lex}) from 38\% to 22.4\%. 61\% of the sentences where one or more assignation error was corrected by the MG module actually have better parsing results (15\% have been degraded). More generally, the increase of guessed forms makes the category assignation more reliable.    

The \textbf{extension of the lexicon} and the \textbf{normalization of genes and species names} discharged the two modules from 143 assignations out of 268, 50 of which were wrong. 64\% of the sentences where one or more assignation error was corrected by  the extension of lexicon have better parsing results (18\% of the sentences were degraded).


\begin{table}
\begin{center}
\begin{tabular}{|c||c||c|c||c|c|}
\hline
 & \footnotesize{lp} & \multicolumn{2}{|c||}{\footnotesize{lp-bio}} &  \multicolumn{2}{|c|}{\footnotesize{lp-bio-t}}\\
\hline
\footnotesize{crit.} &\footnotesize{avg} & \footnotesize{avg} & \footnotesize{\%/lp} & \footnotesize{avg} & \footnotesize{\%/lp} \\
\hline
\hline
\footnotesize{NbW} &\scriptsize{24.05} &\scriptsize{24.05} & \scriptsize{100\%} & \scriptsize{18.9} & \scriptsize{78.6\%}\\
\hline
\footnotesize{NbL} &\scriptsize{190,306} & \scriptsize{232,622} & \scriptsize{122.2\%} & \scriptsize{1,431} & \scriptsize{0.75\%} \\
\hline
\footnotesize{PT} &\scriptsize{37.83} &\scriptsize{29.4} & \scriptsize{77.7\%} &\scriptsize{0.53} & \scriptsize{1.4\%} \\
\hline
\footnotesize{CLF} &\scriptsize{0.54} & \scriptsize{0.72}& \scriptsize{133\%} &\scriptsize{0.77} & \scriptsize{142.6\%}\\
\hline
\footnotesize{EL} & \scriptsize{2.87} &\scriptsize{1.91} &\scriptsize{66.5\%} &\scriptsize{1.15} & \scriptsize{40.1\%} \\
\hline
\footnotesize{CQ} &\scriptsize{0.54} & \scriptsize{0.7} & \scriptsize{129.6\%} &\scriptsize{0.8} & \scriptsize{148.1\%} \\
\hline

\end{tabular}\caption{Parsing time and quality} \label{res}
\end{center}
\end{table}
The effect of the \textbf{rules modification and creation} is difficult to evaluate precisely though it is certain to play a part in the parsing improvement, especially the relaxing of constraints on determiners and inserts. 

The most obvious contribution to the better parsing quality is the one of the \textbf{term simplification}. The drastic reduction in parsing time and number of linkages gives an idea of the reduction of complexity. It is not only due to the smaller number of words since the number of erroneous links is reduced of 60\% while the number of words is reduced of only 21.4\%. This confirms previous similar studies that showed a reduction of 40\% of the error rate on the main syntactic relations with a French corpus.

\textbf{Remaining errors} are mainly due to four different phenomena. First, the normalization step, prior to the parsing, needs to be enhanced. 
Concerning LP, there are still  lexicon gaps, wrong class assignations and a still unsatisfactory handling of numerical expressions. 
In addition, and like \cite{Sut95}, we identified a weakness of LP regarding coordination. 
A specific study of the coordination system in LP and in the biological texts may be necessary. Finally, some ambiguous nominal and prepositional attachments still remain in spite of the term simplification. These may be resolved in a post-processing step like in ExtrAns that uses a corpus based approach to retrieve the correct attachment from the different linkages given by LP for a sentence.

Other questions like the feeding of LP with a morpho-syntactically tagged text or the amelioration of the parse ranking in LP were not discussed in this paper but are interesting issues that we intend to study.

\section{Conclusion}
Since parsing is domain and language dependent, a general parser must be adapted to each given sublanguage. In the context of an IE project in biology, we have adapted the Link Parser to analyse the specific language of Medline abstracts in genomics. Our initial diagnosis mainly raised two different problems which are traditional in sublanguage analysis: the lack of lexical coverage 
and the structural ambiguity, especially in the cases of prepositional phrase attachments.

We showed that the lexical problem can be manually handled by introducing new words in the lexicon and by extending the morpho-guessing module. We also proposed to distinguish and combine terminological and syntactic analysis. In the same way as the morpho-syntactic tagging should be considered independently from the parsing, we argue that the terminology analysis must be handled separately. This represents the main automated part of the adaptation task. 
The use of terminology to alleviate the parsing task is relevant and applicable in the context of domain specific texts processing since terminology tools and lists of terms are generally available. It also reduces the part of effective modification of the lexicon/grammar of the parser. 
This first evaluation has shown promising results.

This work has been developed as part of the ExtraPloDocs (extraction of gene-protein interactions in Medline abstracts) and ALVIS projects. We have shown that combining the terminological and syntactic analysis has an important impact on the resulting parses because the terminological analysis simplifies the parser input.

 \section{Bibliography}
 \label{sec:Bibliography}
 \bibliographystyle{ranlp}
 \begin{scriptsize}
\bibliography{/users/aubin/biblio/latex/sophieAubin}
 \end{scriptsize}

\end{document}